# Implementations in Machine Ethics: A Survey

SUZANNE TOLMEIJER, MARKUS KNEER, CRISTINA SARASUA, MARKUS CHRISTEN,
and ABRAHAM BERNSTEIN, University of Zürich

Increasingly complex and autonomous systems require machine ethics to maximize the benefits and minimize the risks to society arising from the new technology. It is challenging to decide which type of ethical theory to employ and how to implement it effectively. This survey provides a threefold contribution. First, it introduces a trimorphic taxonomy to analyze machine ethics implementations with respect to their object (ethical theories), as well as their nontechnical and technical aspects. Second, an exhaustive selection and description of relevant works is presented. Third, applying the new taxonomy to the selected works, dominant research patterns, and lessons for the field are identified, and future directions for research are suggested.

## 1 INTRODUCTION

Autonomous machines are increasingly taking over human tasks. Initially, simple and limited assignments such as assembly line labor were taken over by machines. Nowadays, more complex tasks are transferred to software and robots. Even parts of jobs that were previously deemed purely human occupations, such as being a driver, credit line assessor, medical doctor, or soldier are progressively carried out by machines (e.g., References [39, 45]). As many believe, ceding control over important decisions to machines requires that they act in morally appropriate ways. Or, as Picard puts it, "the greater the freedom of a machine, the more it will need moral standards" [106, p. 134].

For this reason, there has been a growing interest in *Machine Ethics*, defined as the discipline "concerned with the consequences of machine behavior toward human users and other

This work is partially funded by armasuisse Science and Technology (S+T), via the Swiss Center for Drones and Robotics of the Department of Defense, Civil Protection and Sport (DDPS). Research on this paper was also supported by an SNSF Ambizione grant for the project Reading Guilty Minds (PI Markus Kneer, PZ00P1_179912).

Authors' addresses: S. Tolmeijer, C. Sarasua, and A. Bernstein, Department of Informatics, University of Zurich, Binzmühlestrasse 14, 8050 Zürich, Switzerland; emails: {tolmeijer, sarasua, bernstein}@ifi.uzh.c; M. Kneer, Centre for Ethics, University of Zurich, Zollikerstrasse 118, 8008 Zurich, Switzerland; email: markus.kneer@uzh.ch; M. Christen: Institute of Biomedical Ethics and History of Medicine and Digital Society Initiative, University of Zurich, Winterthurerstrasse 30, 8006 Zürich, Switzerland; email: christen@ethik.uzh.ch.

machines" [6, p. 1].[1] Research in this field is a combination of computer science and moral philosophy. As a result, publications range from theoretical essays on what a machine can or should do (e.g., References [26, 49, 61, 127]) to prototypes implementing ethics in a system (e.g., References [3, 147]). In this field, the emphasis lies on how to design and build a machine such that it could act ethically in an autonomous fashion.[2]

The need that complex machines should interact with humans in an ethical way is undisputed; but for understanding which design requirements follow from this necessity requires a systematic approach that is usually based on a taxonomy. There have been several attempts to classify current approaches of machine ethics. A first high-level classification was proposed by Allen et al. [2] in 2005, distinguishing between top-down theory-driven approaches, bottom-up learning approaches, and hybrids of the two. Subsequent work tried to further determine types of procedures [32, 152, 153], but these works were either mixing different dimensions (e.g., mixing technical approach and ethical theory in one category) [152] or offering an orthogonal dimension that did not fit the existing taxonomy (e.g., whether normative premises can differ between ethical machines) [32]. Also, because these works did not provide an extensive and systematic overview of the application of their taxonomy, verification of the taxonomy with papers from the field was missing. A recent survey from Yu et al. [153] on ethics in AI has some overlap with this work but (1) does not systematically apply the ethical theory classification to selected papers and (2) takes a broader perspective to include consequences of and interaction with ethical AI, while this article focuses specifically on machine ethics implementations. Hence, compared to previous works, this survey covers more related work, provides a more extensive classification, and describes the relationship between different ethics approaches and different technology solutions in more depth than previous work [2, 32, 152, 153]. Furthermore, gaps are identified regarding nontechnical aspects when implementing ethics in existing systems.

This article is created as a collaboration between ethicists and computer scientists. In the context of implementing machine ethics, it can be a pitfall for philosophers to use a purely theoretical approach without consulting computer scientists, as this can result in theories that are too abstract to be implemented. Conversely, computer scientists may implement a faulty interpretation of an ethical theory if they do not consult a philosopher. In such an interdisciplinary field, it is crucial to have a balanced cooperation between the different fields involved.

The contributions of this article are as follows:

- Based on previous work [2], a trimorphic taxonomy is defined to analyze the field based on three different dimensions: types of ethical theory (Section 4), nontechnical aspects when implementing those theories (Section 5), and technological details (Section 6).
- The reviewed publications are classified and research patterns and challenges are identified (Section 7).
- An exhaustive selection and description of relevant contributions related to machine ethics implementations is presented (Appendix A).
- A number of general lessons for the field are discussed and further important research directions for machine ethics are outlined (Section 8).

---

[1]While there are other terms for the field, such as "Artificial Morality" and "Computational Ethics," the term "Machine Ethics" will be used throughout this survey to indicate the field.

[2]In the following, the expression "implementations in machine ethics" concerns all relevant aspects to successfully create real-world machines that can act ethically—namely the object of implementation (the ethical theory), as well as nontechnical and technical implementation aspects when integrating those theories into machines. By "machine" we denote both software and embodied information systems (such as robots).

As such, this survey aims to provide a guide, not only to researchers but also to those interested in the state of the art in machine ethics, as well as seed a discussion on what is preferred and accepted in society, and how machine ethics should be implemented.

The rest of this article is structured as follows. Section 2 introduces the field of machine ethics, its importance, and justification of used terminology throughout the article. Section 3 lists the methodology used to create this survey, including the search methodology, the process of creating the classification dimensions, and the actual classification process. Sections 4, 5, and 6 introduce the three classification dimensions presented in this survey. Section 7 discusses the results of the classification of the selected papers. Finally, Section 8 outlines which future avenues of research may be interesting to pursue based on the analysis, as well as the limitations of this survey.

## 2 INTRODUCTION TO MACHINE ETHICS

Before going into more depth on the implementation of ethics, it is important to establish what is considered machines ethics, why it matters, and present the relevant terminology for the field.

### 2.1 Relevance

Software and hardware (combined under the term "machine" throughout this survey) are increasingly assisting humans in various domains. They are also tasked with many types of decisions and activities previously performed by humans. Hence, there will be a tighter interaction between humans and machines, leading to the risk of less meaningful human control and an increased number of decision made by machines. As such, ethics needs to be a factor in decision making to consider fundamental problems such as the attribution of responsibility (e.g., Reference [127]) or what counts as morally right or wrong in the first place (e.g., Reference [140]). Additionally, ethics is needed to reduce the chance of negative results for humans and/or to mitigate the negative effects machines can cause.

Authors in the field give different reasons for studying (implementations in) machine ethics. Fears of the negative consequences of AI motivate the first category of reasons: creating machines that do not have a negative societal impact [13, 89]. With further autonomy and complexity of machines, ethics need to be implemented in a more elaborate way [27, 41, 52, 78, 98, 102]. Society needs to be able to rely on machines to act ethically when they gain autonomy [6, 51]. A second category of reasons for studying machine ethics focuses on the ethics part: By implementing ethics, ethical theory will be better understood [27, 62, 98, 102]. Robots might even outperform humans in terms of ethical behavior at some point [4, 9].

Some authors contend that in cases with no consensus on the most ethical way to act, the machine should not be allowed to act autonomously [5, 127]. However, not acting does not imply the moral conundrum is avoided. In fact, the decision *not* to act also has a moral dimension [58, 81, 149]—think, for example, of the difference between active and passive euthanasia [111]. Additionally, by not allowing the machine to act, all the possible advantages of these machines are foregone. Take, for example, autonomous cars: A large number of traffic accidents could be avoided by allowing autonomous cars on the road. Moreover, simply not allowing certain machines would not stimulate the conversation on how to solve the lack of consensus, a conversation that can lead to new, more practical ethical insights and helpful machines.

### 2.2 Terminology

An often-used term in the field of machine ethics is "Artificial Moral Agent" or AMA, to refer to a machine with ethics as part of its programming. However, to see whether this term is appropriate to use, it is important to identify what moral agents mean in the context of machine ethics and how ethical machines should be regarded. In an often-cited paper, Moor [98] defines four different levels of moral agents:

**Ethical-impact agents** are types of agents that have an (indirect) ethical impact. An example would be a simple assembly line robot that replaces a human in a task. The robot itself does not do anything (un)ethical by acting. However, by existing and doing its task, it has an ethical impact on its environment; in this case, the human that performed the task is replaced and has to find another job.

**Implicit ethical agents** do not have any ethics explicitly added in their software. They are considered implicitly ethical, because their design involves safety or critical reliability concerns. For example, autopilots in airplanes should let passengers arrive safely and on time.

**Explicit ethical agents** draw on ethical knowledge or reasoning that they use in their decision process. They are explicitly ethical, since normative premises can be found directly in their programming or reasoning process.

**Fully ethical agents** can make explicit judgments and are able to justify these judgments. Currently, humans are the only agents considered to be full ethical agents, partially because they have consciousness, free will, and intentionality.

While these definitions can help with a first indication of the types of ethical machines, they do not allow for distinctions from a technical perspective and are also unclear from a philosophical perspective: Moor [98] does not actually define what a moral agent is. For example, it can be debated whether an autopilot is an agent. Therefore, a clearer definition is needed of what an agent is. Himma [77] investigates the concepts of agency and moral agency, drawing from philosophical sources such as the Stanford Encyclopedia of Philosophy and Routledge Encyclopedia of Philosophy. He proposes the following definitions:

**Agent:** "X is an agent if and only if X can instantiate intentional mental states capable of performing actions" [77, p. 21].

**Moral agency:** "For all X, X is a moral agent if and only if X is (1) an agent having the capacities for (2) making free choices, (3) deliberating about what one ought to do, and (4) understanding and applying moral rules correctly in the paradigm cases" [77, p. 24].

With regards to artificial agents, Himma postulates that the existence of natural agents can be explained by biological analysis, while artificial agents are created by "intentional agents out of pre-existing materials" [77, p. 24]. He emphasizes that natural and artificial agents are not mutually exclusive (e.g., a clone of a living being). He further claims that moral agents need to have conscious and intentionality, something that state-of-the-art systems do not seem to instantiate. It is worth noting that Himma attempts to provide a general definition of moral agency, while, for example, Floridi and Sanders [57] propose to change the current description of a moral agent. For example, they proposed description includes the separation the technical concepts of moral responsibility and moral accountability, a distinction that was not evident thus far: "An agent is morally accountable for x if the agent is the source of x and x is morally qualifiable [...] To be also morally responsible for x, the agent needs to show the right intentional states." Wallach and Allen [142] rate AMAs along two dimensions: how sensitive systems are to moral considerations and how autonomous they are. Sullins [131] has a partially overlapping concept of requirements for robotic moral agency with Himma's that intersects with Wallach and Allen's relevant concepts: autonomy (i.e., "the capacity for self-government") [30]), intentionality (i.e., "the directedness or 'aboutness' of many, if not all, conscious states"[30]), and responsibility (i.e., "those things for which people are accountable"[30]).

These are just some notions of how concepts such as agency, autonomy, intentionality, accountability, and responsibility are important to the field of machine ethics. However, it is challenging to summarize and define these concepts concisely while doing justice to the work in philosophy and computer science that has been done so far, including the discussions and controversy around different relevant concepts (as the different concepts of moral agency display). The goal of this survey is not to give an introduction to moral philosophy, so this section merely gives a glimpse of the depths of the topic. Rather, the goal is to summarize and analyze the current state of the field of machine ethics. To avoid any assumption on concepts, the popular term Autonomous Moral Agent is not used in this survey: As shown above, the term "agent" can be debated in this context and the term "autonomous" has various meanings in the different surveyed systems. Instead, a machine that has some form of ethical theory implemented—implicitly or explicitly—in it is referred to as an "ethical machine" throughout this article. Accordingly, we refrained from adding an analysis regarding degree of agency and autonomy of machines into our taxonomy, as those points are rarely discussed by the authors themselves and because they would have added a layer of complexity that would have made our taxonomy confusing.

## 3 SURVEY METHODOLOGY

This section describes the search strategy, paper selection criteria, and review process used for this survey.

### 3.1 Search Strategy

A literature review was conducted to create an overview of the different implementations of and approaches to machine ethics. The search of relevant papers was conducted in two phases: automated search and manual search.

*Automated Search.* The first phase used a search entry that reflected different terms related to machine ethics combined with the word "implementation":

> implementation AND ("machine ethics" OR "artificial morality" OR "machine morality" OR "computational ethics" OR "roboethics" OR "robot ethics" OR "artificial moral agents")

These terms were cumulated during the search process (e.g., Reference [144, p. 455]); each added term resulted in a new search until no new terms emerged.[3] No time period of publication was specified, to include as many items as possible.

The following library databases were consulted (with the number of results in parenthesis): Web of Science (18), Scopus (237), ACM Digital Library (16), Wiley Online Library (23), ScienceDirect (48), AAAI Publications (4), Springer Link (247), and IEEE Xplore (113). Of these initial results, 37 items were selected based on the selection criteria listed in Section 3.2.

*Manual Search.* The second phase included checking the related work and other work by the same first authors of phase one. Twenty-nine promising results from phase one did not meet all criteria but were included in the second search phase to see if related publications did meet all criteria. This process was repeated for each newly found paper until no more papers could be added that fit the selection criteria (see Section 3.2). This resulted in a total of 49 papers, describing 48 ethical machines.

---

[3]The term "Friendly AI," coined by Yampolsky [154], is excluded, since it describes theoretical approaches to machine ethics.

## 3.2 Selection Criteria

After the selection process, two more coauthors judged which papers should be in- or excluded to verify the selection. Papers were included only if they adhered to all of the following inclusion criteria. The article

- implements a system OR describes a system in sufficient (high-level) detail for implementation OR implements/describes a language to implement ethical cases,
- describes a system that is explicitly sensitive to ethical variables (as described in Reference [98]), no matter whether it achieves this sensitivity through top-down rule-based approaches or bottom-up data-driven approaches (as described in Reference [2]),
- is published as a conference paper, workshop paper, journal article, book chapter, or technical report,
- and has ethical behavior as the main focus.

The following exclusion criteria were used. The article

- describes machine ethics in a purely theoretical fashion,
- describes a model of (human) moral decision making without an implementable model description,
- lists results of human judgment on ethical decisions without using the data in an implementation,
- is published as a complete book, presentation slides, editorial, thesis, or has not been published,
- describes a particular system in less detail than other available publications,
- focuses on unethical behavior to explore ethics (e.g., a lying program),
- mentions ethical considerations while implementing a machine, but does not focus on the ethical component and does not explain it in enough detail to be the main focus,
- simulates artificial agents to see how ethics emerge (e.g., by using an evolutionary algorithm without any validation),
- and describes a general proposal of an ethical machine without mentioning implementation related details.

Given the focus on algorithms implementing moral decision making and the limitations of space, we will not go into further detail as regards recent interesting work on AI and moral psychology (cf. e.g., References [18, 31, 88, 123]).

## 3.3 Taxonomy Creation and Review Process

To be able to identify strengths and weaknesses of the state of the art, we created different taxonomies and classified the selected papers accordingly. It was clear that both a dimension referring to the implementation object (the ethical theory; cf. Table 1) and a dimension regarding the technical aspects of implementing those theories (cf. Table 4) were necessary. All authors agreed that there were some aspects of implementing those theories that did not concern purely technical issues but were still important for the classification. Hence, we defined and applied a third taxonomy dimension (cf. Table 3) related to non-technical implementation choices. The first version of these three taxonomies was created using knowledge obtained during the paper selection.

Before the classification process started, one-third of the papers were randomly selected to review the applicability of the taxonomy proposed and adjust the assessment scheme where required. Any parts of the taxonomies that was unclear and led to inconsistent classifications was adapted and clarified.

Each selected paper was categorized according to the features of the three different taxonomy dimensions (discussed in Sections 4–6). Since the ethical classification is perhaps the most disputable, it was determined by three distinct assessors: two philosophers and a computer scientist. Two computer scientists evaluated the implementation and technical details of all proposed systems.

To provide a classification for all selected papers, multiple classification rounds took place for each dimension. Between classification rounds, disagreements across reviewers were discussed until a consensus was reached. In the case of the ethical dimensions, four papers could not be agreed upon after multiple classification rounds. As such, these papers were labeled as "Ambiguous."

Additionally, we shared a pre-print of the article with the authors of the classified systems to verify that they agreed with the classifications we provided. From 45 targeted authors, we received 18 responses. From these 18, 6 authors agreed with our classification and 12 proposed (mostly minor) changes or additional citations. In total, we changed the classification of 6 features for 4 papers.

In the following, we now outline the three dimensions of our taxonomy.

## 4 OBJECT OF IMPLEMENTATION: ETHICAL THEORIES

This section introduces the first of three taxonomy dimensions introduced in this article: a taxonomy of types of ethical theories, which is the basis for the categorization of ethical frameworks used by machines (in Section 7). Note that this section is not a general introduction to (meta-)ethics, which can for example be found in References [29, 44, 97].

### 4.1 Overview of Ethical Theory Types

It is commonplace to differentiate between three distinct overarching approaches to ethics: *consequentialism*, *deontological ethics*, and *virtue ethics*. Consequentialists define an action as morally good if it maximizes well-being or utility. Deontologists define an action as morally good if it is in line with certain applicable moral rules or duties. Virtue ethicists define an action as morally good if, in acting in a particular way, the agent manifests moral virtues. Consider an example: An elderly gentleman is harassed by a group of cocky teenagers on the subway and a resolute woman comes to his aid. The consequentialist will explain her action as good, since the woman maximized the overall well-being of all parties involved—the elderly gentleman is spared pain and humiliation, which outweighs the teenagers' amusement. The deontologist will consider her action commendable as it is in accordance with the rule (or duty) to help those in distress. The virtue ethicist, instead, will deem her action morally appropriate, since it instantiates the virtues of benevolence and courage.

Consequentialist theories can be divided into two main schools: According to *act utilitarianism*, the principle of utility (maximize overall well-being) must be applied to each individual act. *Rule utilitarians*, by contrast, advocate the adoption of those and only those moral rules that will maximize well-being. Cases can thus arise where an individual action does not itself maximize well-being yet is consistent with an overarching well-being maximizing rule. While act utilitarians would consider this action morally bad, rule utilitarians would consider it good.

Deontological ethics can be divided into agent-centered and patient-centered approaches. *Agent-centred theories* focus on agent-relative duties, such as, for instance, the kinds of duties someone has toward their parents (rather than parents in general). Theories of this sort contrast with *patient-centered theories* that focus on the rights of patients (or potential victims), such as the right, postulated by Kant, not to be used as a means to an end by someone else [84].

Finally, there are some approaches that question the universal applicability of general ethical principles to all situations, as put forward by deontological ethics, virtue ethics, or consequentialism. For such a *particularist view*, moral rules or maxims are simply vague rules of thumb, which cannot do justice to the complexity of the myriad of real-life situations in which moral agents might find themselves. Hence, they have to be evaluated on a case-by-case basis.

Table 1. Ethical Theory Types Taxonomy
Dimension

| Ethics Type |
| --- |
| Deontological ethics |
| Consequentialism |
| Virtue ethics |
| Particularism |
| Hybrid |
| — *Hierarchically specific* |
| — *Hierarchically nonspecific* |
| Configurable ethics |
| Ambiguous |

We highlight that a moral theory is a set of substantial moral principles that determine what, according to the theory, is morally right and wrong. Moral theories can take different structures—they might state their concrete demands in terms of hard rules (deontological ethics); virtues that should guide actions, with reference to an overall principle of utility maximization, or else reject the proposal that there is a one-size-fits-all solution (itself a structural trait, this would be particularism). In this work, we are interested in these structures, which we label "ethical theory types."

## 4.2 Categorizing Ethical Machines by Ethical Theory Type

Based on the distinct types of ethical theories introduced above, this sub-section develops a simple typology of ethical machines, summarized in Table 1.

An evaluation of existing approaches to moral decision making in machines can make use of this typology in the following way. Deontological ethics is rule based. What matters is that the agent acts in accordance with established moral rules and/or does not violate the rights of others (whose protection is codified by specified rules). Accidents occur, and a well-disposed agent might nonetheless bring about a harmful outcome. On off-the-shelf deontological views, bad outcomes (if non-negligently, or at least unintentionally, brought about) play no role in moral evaluation, whereas the agent's mental states (their intentions and beliefs) are important. If John, intending to deceive Sally about the shortest way to work, tells the truth (perhaps because he himself is poorly informed), then a Kantian will consider his action morally wrong, despite its positive consequence.[4] In the context of machine ethics, the focus is solely on agent relative duties. Hence, no distinction is made between agent-centered and patient-centered theories of deontological ethics in the taxonomy summarized in Table 1.

Consequentialists, by contrast, largely disregard the agent's mental states and focus principally on outcomes: What matters is the maximization of overall well-being. Note that, procedurally, a rule-utilitarian system can appear very similar to a deontological one. The agent must act in keeping with a set of rules (potentially the very same as in a Kantian system) that, in the long run, maximizes well-being. However, the two types of systems can still be distinguished in terms of the ultimate source of normativity (well-being vs. good will) and will—standardly—differ in terms of the importance accorded to the agent's mental states. Thus far, nearly all consequentialist machine ethics implementations utilize act utilitarianism. For this reason, the distinction between act and rule utilitarianism is not relevant enough to be included in this survey.

---

[4]Note that if two actions differ only with respect to outcome, then consequences can play a role.

Table 2. High-level Overview to Ethics Categories in the Context of Ethical Machine Implementation

|  | Input | Decision criteria | Mechanism | Challenges (examples) |
|---|---|---|---|---|
| **Deontological ethics** | Action (mental states and consequences) | Rules/duties | Fittingness with rule | • Conflicting rules<br>• Imprecise rules |
| **Consequentialism** | Action (consequences) | Comparative well-being | Maximization of utility | • Aggregation problems<br>• Determining utility |
| **Virtue ethics** | Properties of agent | Virtues | Instantiation of virtue(s) | • Conflicting virtues<br>• Concretion of virtues |
| **Particularism** | Situation (context, features, intentions, consequences) | Rules of thumb, precedent, all situations are unique | Fittingness with rules/precedent | • No unique and universal logic<br>• Each situation needs unique assessment |

Virtue ethics differs from the aforementioned systems insofar as it does not focus principally on (the consequences or rule-consistency of) actions but on agents and more particularly on whether they exhibit good moral character or virtuous dispositions. A good action is one that is consistent with the kinds of moral dispositions a virtuous person would have.

In contrast to the other three major approaches, on the particularist view, there is no unique source of normative value, nor is there a single, universally applicable procedure for moral assessment. Rules or precedents can guide our evaluative practices. However, they are deemed too crude to do justice to many individual situations. Thus, according to particularism, whether a certain feature is morally relevant or not in a new situation—and if so, what exact role it is playing there—will be sensitive to other features of the situation.

Table 2 gives a schematic overview of key characteristics of the different types of ethical systems that might be implemented in an ethical machine. Note that it does not take some of the more fine-grained aspects differentiating the theories (e.g., the before-mentioned complications regarding act and rule utilitarianism) into account.

As an alternative to implementing a single determinate type of ethics, systems can also combine two or more types, resulting in a *hybrid* ethical machine. This approach seems enticing when one theory alleviates problems another one might have in certain situations, but it can also generate conflicts across types of ethical approaches. Hence, some proposals enforce a *specified hierarchy*, which means that one theory is dominant over the other(s) in the system. For example, a primarily deontological system might use rules but turn to the utilitarian approach of maximizing utility when the rules are in conflict. In other cases, the hierarchy is *non-specific* and different theories are present without a specified dominant theory.

Some authors do not settle on a particular type of ethical theory. Instead, they provide a *configurable* technical framework or language and exhibit how different types of ethical theories can be implemented. The choice of which theory type should be selected is essentially left to the person implementing the system in an actual use case.

Finally, some contributions were classified as *ambiguous* from a meta-ethical perspective. For these, not enough details were given by the authors to classify a paper or the theories used to implement were not ethical theories but retrieved from domains other than moral philosophy.

## 4.3 Ethical Theory Types in Practice

There are certain challenges inherent in the different types of ethics when they need to be applied in practice. Since these obstacles need to be taken into account to select an ethical theory type for an ethical machine, this subsection provides a (non-exhaustive) list of complications.

*Challenges of deontological ethics in practice:* At a first glance, the rule-based nature of deontological ethics seems to lend itself well for implementation. However, at different stages of implementation, challenges arise. The first issue is *which rules should be implemented*. Rules are expected to be strictly followed, implying that for every exception, the rule must be amended, resulting in an extremely long rule. Determining the right level of detail is important for the success of an application: when the rules are not practical and at the right level of detail, they will not be interpretable for the machine [9]. Second, there might be *conflicts between rules* [32]—in general or in specific situations. Whilst ordering or weighing the rules might address this issue from an implementational perspective, determining an order of importance can be difficult. Also, this assumes that all relevant rules are determined before they are used.

*Challenges of consequentialist ethics in practice:* There are three main categories of difficulties for consequentialist ethics. First, it is hard to identify consequences and determine the right level of detail and aggregation in terms of time and size. Some outcomes might have resulted regardless of the action theorized to have caused it. In real-life situations, all possible consequences are not always that clear beforehand given the lack of epistemic transparency and causal interdependence.

A second issue is concerned with quantifying consequences. As consequentialism is about maximizing utility, the problem is how to define *utility*. In simple scenarios like the Trolley problem, utility is often defined as how many people survive or die. In the real world, more complex concepts, such as happiness and well-being, are preferred to define utility. There are measures available (e.g., QALY [68]), but using a different measure can give a different outcome. Even more so, even if each consequence is assigned a utility, then it might still be inappropriate to simply aggregate them (e.g., see Reference [86]).

Finally, there might be a significant computational cost when computing utility [146] requiring heuristics or approximations to derive a correct answer in time. This, in turn, requires a verification of whether these results are still correct.

*Challenges of virtue ethics in practice:* Virtues are positive character traits, character traits that should be manifested in morally good actions. Defining what "character" a machine has is troubling, if a machine can be claimed to have a character at all. To judge whether a machine—or a human for that matter—is virtuous is not possible by merely observing one action or a series of actions that seem to imply that virtue; the reasons behind them need to be clear [128]. Perhaps the best way to create a virtuous machine is to let a machine mimic the behavior of a virtuous person. But how is a certain virtue measured, and who decides which virtues are more important and how to pick the perfect role model? Coleman [42] even proposes different virtues that are more desirable for machines rather than human virtues, implying merely mimicking a virtuous person is not sufficient.

To circumvent these challenges, machine ethics researchers have not used virtue ethics often, as the alternatives might be more appealing. For example, Haber [69] states that virtue ethics and principle-based ethics are complements and that for each trait there will be a principle that expresses that trait and vice versa. While not everyone agrees with Haber, it is easier and more detailed from a computational perspective to implement rules than generic virtues to adhere to. Arkin [9] also concludes that principle-based and act-centric models allow for stricter ethical implementations, which is desirable in machine ethics.

*Challenges of particularism in practice:* In particularism, the system needs to take the entire context into account. This implies that it needs to either be trained for all possible cases, which is not possible, or be able to extrapolate without using generalizations, which is highly challenging. For each feature of the context, the system would have to recognize whether it is morally relevant in the given case and how it will influence the result. Case-based methods or instance-based classifications come closest to allowing an implementation of particularism. More recently, some contributions are trying to approximate particularist ethics using neural networks (e.g., References [66, 74]).

Table 3. Non-technical Taxonomy Dimension

| Feature | Type | Subtype |
|---|---|---|
| **Approach** | Top-down | |
| | Bottom-up | |
| | Hybrid | |
| **Diversity consideration** | Yes | |
| | No | |
| **Contribution type** | Model representation | |
| | Model selection | |
| | Judgment provision | |
| | Action selection/execution | |
| **Evaluation** | Test | Non-expert |
| | | Expert |
| | | Laws |
| | Prove | Model checker |
| | | Logical proof |
| | Informal | Example scenarios |
| | | Face validity |
| | None | |
| **Domain specific** | Yes (*domain specified*) | |
| | No | |

*Challenges of hybrid approaches in practice:* Each type of ethical theory raises its own set of complications, but combining them introduces additional issues. First, when different types of ethical theories are used in a non-hierarchical way, the interaction between them can be problematic: How should the results from different ethical approaches be combined to guarantee morally appropriate outcomes? What happens when the results of different implemented ethical theory types stand in conflict, and how should such conflicts be resolved?

Second, when a hierarchical approach is employed, it is not evident when the system should employ one theory rather than another. One standard approach resorts to the secondary set of ethical principles when the first does not deliver a verdict. While this alleviates some of the challenges of hybrid systems, it is still possible that the second ethical theory proposes something that conflicts with the first ethical theory type.

The next section introduces the second dimension of ethical machines: the non-technical aspects of implementing ethics into a machine.

## 5 NON-TECHNICAL IMPLEMENTATION ASPECTS

The second taxonomy dimension that was created for this survey considers the nontechnical aspects of implementing the aforementioned ethical theories into machines. An important part of creating an ethical system is to decide *how* to implement ethics. That entails defining whether an implementation can follow different approaches, how to evaluate the system, and whether or not domain specifications need to be taken into account. Important features concerning the implementation dimension are summarized in Table 3. Furthermore, this section highlights the implementation challenges that the various ethical theories entail.

## 5.1 Approaches

Different typologies have been proposed to determine how ethics types are implemented. The most influential and widely referenced scheme, also applied in this survey, stems from Allen, Smit, and Wallach [2]. They distinguish three types of implementation approaches, namely top-down, bottom-up, and hybrid.

**Top-down approaches:** Top-down approaches assume that humans have gathered sufficient knowledge on a specific topic; it is a matter of translating this knowledge into an implementation. The ethical theory types described in Section 4 are examples of normative human knowledge that can be translated into usable mechanisms for machines. The system acts in line with predetermined guidelines and its behavior is therefore predictable. In AI, strategies using a top-down approach mostly make use of logical or case-based reasoning. Given general domain knowledge, the system can reason about the situation that is given as input. Usually, human knowledge is not specified in a very structured or detailed way for concrete cases, so knowledge needs to be interpreted before it can be used. This process presents the risk of losing or misrepresenting information. The positive aspect of this approach is that existing knowledge is applied and no new knowledge needs to be generated.

**Bottom-up approaches:** A different method to implementing ethics is to assume the machine can learn how to act if it receives as input enough correctly labeled data to learn from. This approach, not just in machine ethics but in general, has gained popularity after the surge of machine learning in AI and the recent success of neural networks. Technologies such as artificial neural networks, reinforcement learning, and evolutionary computing fall under this trend. Increased computing power and amounts of data allow learning systems to become more successful. However, data has to be labeled consistently and the right data properties need to be described in a machine-processable way to obtain an accurate training of machines. There is a risk that the machine learns the wrong rules or cannot reliably extrapolate to cases that were not reflected in its training data. However, for certain tasks, such as feature selection or classification, machine learning can be very successful.

**Hybrid approaches:** As the term suggests, hybrid approaches combine top-down and bottom-up approaches. As Allen et al. phrase it: "Both top-down and bottom-up approaches embody different aspects of what we commonly consider a sophisticated moral sensibility" [2, p. 153]. They indicate that a hybrid approach is considered necessary, if a single approach does not cover all requirements of machine ethics. The challenge consists in appropriately combining features of top-down and bottom-up approaches.

Bonnemains et al. [32] suggest adding a fourth category, called "Personal values/ethics system." Essentially, it acknowledges that two different agents may rely on different ethical systems or may rely on different precedence in case of conflicts in a hybrid system. In this survey, this is regarded as *diversity consideration*: The authors of a machine ethics paper consider the possibility that not all ethical machines adhere to the same ethical theory type, and their contribution includes the choice of diverse types of ethics to be implemented. As Bonnemains et al. recognize, this category is somewhat orthogonal to the previous three, as all of those can be seen to implement distinct normative principles. For example, a machine ethics implementation with diversity consideration could allow for multiple ethical theory types to be implemented (i.e., a top-down approach) or allow for different machines to learn different types of ethics (i.e., a bottom-up approach). It is considered part of the implementation dimension rather than the ethics dimension, since diversity considerations can also exist within the same ethical theory, for example, by allowing deontological machines to have different rules to adhere to while still all being deontological in nature. This survey regards structures of normative frameworks and their implementation rather than substantial normative principles (cf. Table 3).

## 5.2 Type of Contribution

Ethical systems can be intended to enact different aspects of ethical behavior. This section discusses the different types of contributions published to implement ethical machines.

**Model representation:** This contribution type focuses on representing current ethical knowledge. The goal is to determine how to appropriately represent a theory, dilemma, or expert-generated guidelines whilst staying true to the original theory.

**Model selection:** Given a set of alternative options to implement an ethical machine, some systems limit their action to selecting the most fitting elements to be included in the system.

**Judgment provision:** These contributions focus on judging an action given a scenario and a set of possible actions. Example outputs are binary (*acceptable*/*non-acceptable*) or responses on a scale (e.g., *very ethical* to *very unethical*).

**Action selection/execution:** Here the proposed system chooses which action is best given multiple possible actions for a scenario. Some systems then assign the action to a human, while others carry out the selected action themselves. Part of the action selection task can also be action restriction, when some possible actions are not morally acceptable (enough).

## 5.3 Evaluation

Most artifacts—simple or complex, concrete or abstract—can be evaluated in virtue of their capacity to fulfill their constitutive function or purpose. A good knife cuts well, a good thermostat reliably activates the heating if the temperature drops below a predetermined threshold, and a good translation system adequately and idiomatically converts grammatical sentences from one language into another. Whereas there are objective and measurable criteria for the evaluation of thermostats, things are more cumbersome when it comes to moral machines. This is not because their purpose does not standardly consist in simply "acting morally" but in executing certain tasks (taking care of the elderly, counselling suicidal people, evaluating risk of recidivism, etc.) in a moral fashion. Much rather, the complication arises from the question of what exactly is to count as executing the task at hand in morally appropriate ways, or against what exactly the behavior of the system should be evaluated.

There are objective facts as to whether an image represents a certain type of animal or not. These facts constrain whether the image is correctly classified as representing an animal. The existence of objective, universal moral values, by contrast, is controversial (cf. e.g., References [75, 91, 110, 138]). Furthermore, and as objectivists readily acknowledge, delineating what is morally permissible poses an epistemic challenge of a different order than identifying, say, a giraffe in an image, or determining the weight of an object. The ontological and epistemic complications that arise in the moral domain thus make it difficult to settle on standards against which the performance of a moral machine could be evaluated. More fundamentally, it is not even evident what kinds of considerations should guide the process of choosing such standards.

While complications as to the evaluation of a moral machine are worrying, their practical significance should not be exaggerated. Although there is disagreement as regards complex cases, in ordinary life situations in which one is confronted with extremely difficult ethical decisions or run away trolleys are exceedingly rare. In many domains, moral dilemmas are unlikely to arise or be of much import, and there is widespread convergence (not only among the folk, but experts, too) on what constitutes adequate moral behavior. Overall, then, the challenge of evaluation might raise metaphysical and epistemic complications of limited pragmatic importance, at least when care is exercised to limit the decision capacity of moral machines to mundane contexts that steer clear of complex ethical paradoxes.

*5.3.1  Test.* When a system is tested, the system outcome needs to be compared against a ground truth. These may have the following origins:

**Non-experts:** One possibility consists in making folk morality the benchmark. Problematically, there is substantial evidence of moral parochialism across cultures (e.g., References [56, 90, 118]), and it is not difficult to find topics on which a single nation is roughly divided—just think of abortion, euthanasia, or same-sex relations in the U.S. [117]. Furthermore, the existence of widespread convergence in moral opinion does not necessarily make such opinions true or acceptable (consider that until a century and a half ago, there was broad agreement in considerable parts of the world that slavery is morally acceptable).

**Experts:** To escape the tyranny of a potentially mistaken or self-serving majority, one might adopt the standard of experts in normative ethics. Problematically, however, experts themselves are sometimes deeply divided on fundamental issues of moral import as well as meta-ethical intuitions [34], and their very expertise can be called into question [124, 125].

**Laws:** One might side-step the complications raised by retreating to a second-best solution: the law. This strategy, however, is not without drawbacks either, as the law is simply silent on most questions of day-to-day morality. It is, for instance, not illegal to lie in most contexts, yet it would be regarded as outrageous to be perpetually deceived by "moral" machines. Still, it might be suitable to draw on the law to provide restrictions where they exist, for example, as concerns the "Laws of War" or "Laws of Armed Conflicts" for the lethal weapons domain [9], or specific domain rules such as the Code of Ethics for Engineers [94]. As Arkin [9] suggests, scoping the problem using domain-specific requirements can make it more easily implementable and testable.

*5.3.2  Prove.* Another approach, typically based on some type of logic, consists of proving that the system behaves correctly according to some known specifications. This approach can be divided into the following types:

**Model checker:** Given an ethical machine, a model checker exhaustively and automatically ascertains that it adheres to a given set of specifications.

**Logical proof:** This approach provides a logical proof that given certain premises, the system does what it should do. Proofs of this sort can be effected manually or by using a theorem prover that employs automated logical and mathematical reasoning.

Note that this approach assumes that a correct specification exists *a priori* and is widely accepted. Within the logic community, model checking and theorem proving are often considered an implementation issue rather than a type of evaluation (e.g., see Reference [71]). In some cases, authors do not even explicitly mention that they employ a model checker, because it is inherent in their approach to logic programming. However, given the multidisciplinary nature of the field of machine ethics, it is vital to explicitly state which approach has been used. Furthermore, while logical/internal validity and consistency may be inherent in the system, a form of evaluation is necessary to ensure the system acts as expected in different cases and exhibits external validity.

*5.3.3 Informal Evaluation.* Some authors refrain from formally evaluating their implementation. Instead, they only describe their work and, in some cases, show a few example scenarios or exhibit application domains. Whilst these approaches may have limited validity, they may be warranted given the evaluation complications outlined above or when the authors principally engage in theory building [53].

**Example scenarios/case studies:** To showcase that the system works as intended, one or multiple scenarios are presented to demonstrate the system's performance. This procedure gives a first indication of the functionalities of the machine or may help in theorizing about certain properties of a system, but it does not cover all possible situations or give a complete performance indication.

**Face validity:** Often described as "reasonable results," authors using this approach state that the results of a few example tasks are as expected. It is often unclear what this means and to what extent these results are desirable.

*5.3.4 None.* When no evaluation could be discerned, papers were categorized as having none of the evaluation types present.

## 5.4 Domain Specificity

What is deemed an appropriate action can depend on the domain in which the moral agent is operating, such as the principles in the domain of biomedical ethics as proposed by Beauchamps and Childress [22] for the medical domain, or the Rules of Engagement and Laws of Armed Conflict for autonomous weapon systems [9]. Hence, some contributions focus on a specific application domain, which limits the scope of an ethical machine implementation, and thus the endeavor is more manageable [9].

## 6 TECHNICAL IMPLEMENTATION ASPECTS

The third and final taxonomy dimension introduced concerns the technical aspects when implementing ethical theories into machines. This includes the type of technology chosen for the implementation, the input the system relies on, the ethical machine's availability (i.e., implementation details are published), and other technical features: whether it relies on specific hardware or feedback from users, provides explanations for its conclusions, has a user interface (UI), and whether the input for the system needs to be preprocessed. Important features pertaining to the technical dimension are surveyed in Table 4.

### 6.1 Types of Technology

Inspired by Russell and Norvig [115], different types of technologies can be distinguished. While these types of technology are not always clearly delimited, this categorization allows comparing implementations.

*6.1.1 Logical Reasoning.* There are different types of logic or logic-based techniques used in machine ethics.

**Deductive logic:** This is the classical type of logic: Knowledge is represented as logical statements—propositions and rules—that allow deriving new propositions. Pure deductive systems typically involve no learning or inference involved but only derive what can be known from their set of statements and inputs.

Table 4. Technical Taxonomy Dimension

| Feature | Type | Subtype or classification scheme |
|---|---|---|
| | Logical reasoning | Deductive logic |
| | | Non-monotonic logic |
| | | Abductive logic |
| | | Deontic logic |
| | | Rule-based system |
| | | Event calculus |
| | | Knowledge representation & Ontologies |
| | | Inductive logic |
| **Tech type** | Probabilistic reasoning | Bayesian approach |
| | | Markov models |
| | | Statistical inference |
| | Learning | Inductive logic |
| | | Decision tree |
| | | Reinforcement learning |
| | | Neural networks |
| | | Evolutionary computing |
| | Optimization | |
| | Case-based reasoning | |
| **Input** | Case | Logical representation |
| | | Numerical representation |
| | | (Structured) language representation |
| | Sensor data | |
| **Implementation availability** | Specification details | Y - P - N (Yes - Partially - No) |
| | Implementation details | Y - P - N |
| | Code (link) provided | Y - P - N |
| **Other** | Hardware (simulation) | Y - P - N |
| | Feedback | Y - P - N |
| | Explanation | Y - P - N |
| | UI(mostly GUI) | Y - P - N |
| | Automated processing | Y - P - N |

As explained in Section 6, "Inductive logic" is present twice.

**Non-monotonic logic:** Non-monotonic logic allows the revision of conclusions when a conflict arises, for example, in light of new information.

**Abductive logic:** In abductive logic, the conclusions drawn are the most likely propositions given the premises.

**Deontic logic:** This type of logic stems from philosophy and is specifically designed to express normative propositions. Naturally, this type of logic is inherently suited for the representation and deduction of moral propositions.

**Rule-based systems:** As the name suggests, rule-based systems are systems that function based on a set of rules. These can be ethical rules the system has to adhere to. Note that many of the different types of logic above are typically implemented as some form of rule-based system.

**Event calculus:** Event calculus allows reasoning about events. When a machine needs to act ethically, different events can trigger different types of behavior.

**Knowledge representation (KR) and ontologies:** A KR approach focuses on representing knowledge in a form that a computer system can utilize. In other words, the emphasis lies on improving the quality of the data rather than (just) improving the algorithm.

**Inductive logic:** When relying on inductive logic, premises are induced or learned from examples, rather than pre-defined by a human.

*6.1.2 Probabilistic Reasoning.* Recently, probabilistic reasoning has gained more attention. Different types of probabilistic reasoning approaches can be distinguished.

**Bayesian approaches:** Based on Bayes's rule, these approaches rely on prior knowledge to compute the likelihood of an event. In an ethical context, a machine can then act based on this predicted information.

**Markov models:** Markov models focus on sequences of randomly changing events, assuming that a future event only depends on the current (and not the previous) event(s).

**Statistical inference:** By retrieving probability distributions from available data, the system can try to predict the chances of future events happening.

*6.1.3 Learning.* The increased computational power, the amounts of data available, and the GPU-driven revival of neural networks have made learning systems more popular. There are different learning approaches to be characterized.

**Inductive logic:** In inductive logic, a rule-base for reasoning is learned. As such, it is listed under both the "Logic" and "Learning" categories of this taxonomy.

**Decision tree:** Decision trees are a supervised learning method to solve a classification problem by exploring the decision space as a search tree and computing the expected utility. They are, thus, useful to identify and interpret the features that are most important to classify cases.

**Reinforcement learning:** A system can learn from its actions when they are reinforced with rewards or punishments received from its environment.

**Neural networks:** A neural network can be trained on many cases, to be able to classify new cases based on their relevant features.

**Evolutionary computing:** Evolutionary algorithms are used when, for example, different competing models of an ethical machine exist. Models evolve in an iterative fashion, based on actions inspired from the concept of evolution in the field of biology (e.g., selection, and mutation) [82].

*6.1.4 Optimization.* The most common form of optimization relies on a closed-form formula for which some optimal parameters are sought. Different actions get assigned different values based on a predetermined formula, and the best value is chosen (e.g., the highest value).

*6.1.5 Case-based Reasoning.* In case-based reasoning, a new situation is assessed based on a collection of prior cases. Similar cases are identified and their conclusions are transferred to apply to the current situation.

## 6.2 Input

To be able to respond appropriately, ethical machines need to receive information about the environment (or situation at hand). Input is the information that the system receives, not the transformation the system itself performs on the data afterward.

**Sensor data:** In the case of (simulated) hardware, the machine perceives the input through its sensors. The sensor data are interpreted and processed to serve as the input.

**Case: Logical representation:** Systems using a form of logic often need an input case represented using logic.

**Case: Numerical representation:** Other systems, for example ones using neural nets, need their input in a numerical form. This can be a vector representation or a set of numbers.

**Case: Language representation:** Language inputs can be natural language or input translated into structured language.

## 6.3 Implementation Availability

As mentioned before, part of the field of machine ethics tends to be of a theoretical nature. This becomes apparent in the level of detail of implementation proposals. While some authors implement an idea and provide the source code, this is fairly rare in machine ethics. Some authors only give a few implementation details, and others merely specify a high-level description of their idea. Usually, the focus lies on sketching an idea rather than its complete implementation.

**Specification details:** This level has the fewest implementation details: The author specifies the proposed idea (e.g., textually) without any additional detail.

**Implementation details:** This next level provides implementation details illustrating how the specification is implemented in the described machine.

**Code (link) provided:** This final level provides the link to the code of the machine, so the prototype can be used and the experiments can be replicated.

## 6.4 Other Implementation Categories

This section introduces different and independent categories that are of interest for the implementation of an ethical machine.

*6.4.1 Hardware.* Robots can have direct physical results rather than "just" digital or indirect physical consequences. Hardware can change the way people interact with a system and how it should be able to function, making it an interesting and important feature to classify.

*6.4.2 Feedback.* No matter which ethical approach is used, feedback is a valuable component of an ethical system. For example, the user can be asked whether the provided output was the best given the input or whether the system was clear during its decision process.

*6.4.3 Explanation.* Transparency is important when it comes to algorithmic decisions, both from a user perspective [155] and, in some cases (such as the General Data Protection Regulation in the European Union [64]), from a legal perspective. To achieve this goal, an understandable explanation should be provided by the system.

*6.4.4 User Interface.* Systems should be easy to interact with. This is important for all machines, including ethical machines.

*6.4.5 Automated Processing.* Sometimes, initial prototypes focus on the concept of a system, not the (detailed) implementation, and may require some pre-processing of the input data. Ideally, systems should be able to process input from the environment automatically.

## 7 ANALYSIS

The goal of this section is to classify the surveyed moral machines according to the three taxonomy dimensions introduced in Sections 4–6 and elicit patterns in the literature based on this classification. Specifically, every publication is categorized according to the object of implementation (ethical theory) as well as the non-technical and technical aspects of implementing those theories (as described in Section 3). Summaries of the selected papers can be found in Appendix A.

Table 5. Ethical Theory Classification

| Ethical theory type | Papers |
|---|---|
| **Deontological (D)** | Anderson et al. 2004 (W.D.) [6], Anderson et al. 2006 [7], Anderson et al. 2008 [3], Anderson et al. 2014 [5], Bringsjord et al. 2012 [35], Dennis et al. 2016 [51], Malle et al. 2017 [93], McLaren 2003 [94], Mermet et al. 2016 [95], Neto et al. 2011 [100], Noothigattu et al. 2018 [101] Reed et al. 2016 [113], Shim et al. 2017 [126], Turilli 2007 [134], Wiegel et al. 2009 [146] |
| **Consequentialist (C)** | Abel et al. 2016 [1], Anderson et al. 2004 (Jeremy) [6], Armstrong 2015 [14], Cloos 2005 [40], Dennis et al. 2015 [52], Dang et al. 2017 [135], Vanderselst et al. 2018 [137], Winfield et al. 2014 [147], Atkinson et al. 2008 [17] |
| **Particularism (P)** | Ashley et al. 1994 [15], Guarini 2006 [66] |
| **Hybrid dominance D-C** | Arkin 2007 [9], Azad-Manjiri 2014 [19], Dehghani et al. 2008 [50], Govindarajulu et al. 2017 [65], Pereira et al. 2007 [102], Tufis et al. 2015 [133] |
| **Hybrid dominance C-D** | Pontier et al. 2012 [108] |
| **Hybrid undefined dominance** | Lindner et al. 2017 [89] *(C & A)*, Yilmaz et al. 2017 [152] *(D, C & A)*, Honarvar et al. 2009 [78] *(C & P)*, Howard et al. 2017 [82] *(P & Virtue ethics)*, Berreby et al. 2017 [28] *(D & C)* |
| **Configurable ethics** | Bonnemains et al. 2018 [32], Cointe et al. 2016 [41], Ganascia 2007 [62], Thornton et al. 2017 [132] |
| **Ambiguous (A)** | Han et al. 2012 [72], Cervantes et al. 2016 [38], Madl et al. 2015 [92], Verheij et al. 2016 [139], Wallach et al. 2010 [144], Wu et al. 2017 [151], Arkoudas et al. 2005 [13], Furbach et al. 2014 [60] |

*Hybrid dominance D-C* implies both D and C are implemented, but D is dominant. The reverse is true for *Hybrid dominance C-D*. For the *Hybrid undefined dominance* the theories that are combined are noted in parentheses following the citation.

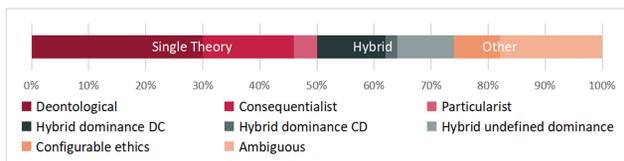

Fig. 1. Ethical theory type ratio.

## 7.1 Ethical Classification

The classification results for the ethical dimension of machine ethics implementations can be found in Table 5; the ratio of single vs. hybrid theory papers is visualized in Figure 1. Among the papers, several constitute clear-cut cases instantiating one of the four main ethical systems. For example, References [5, 100, 126] are clearly deontological, and References [1, 40, 52, 137] constitute uncontroversial examples of consequentialist systems. Furthermore, a considerable number of papers invoke elements from multiple systems. Finally, there are papers in which the hierarchy across theory types remains ambiguous. Examples of ambiguous papers are implementations where authors try to mimic the human brain [37], or focus on implementing constraints such as the Pareto principle [89], which does not strictly speaking constitute a moral theory. Note that categorizing a paper as "ambiguous" does not imply a negative assessment of the implementation. It simply means that the proposal cannot be adequately placed within our classification framework.

About 50% of the proposals draw on a single type of ethical theory (see Figure 1). As can be seen in Table 5, deontological and consequentialist ethics are used most often. It stands out that particularism is barely used and pure virtue ethics is not used at all. This may be explained as follows: First, a generalist approach is much easier to implement than a particularist approach, as it is more straightforward to encode generalist rules than to build systems that may have to handle as of yet unknown, particular cases. Second, virtue ethics can be considered a very high-level theory focusing on characteristics rather than actions or consequences, which is difficult to interpret in an application context.

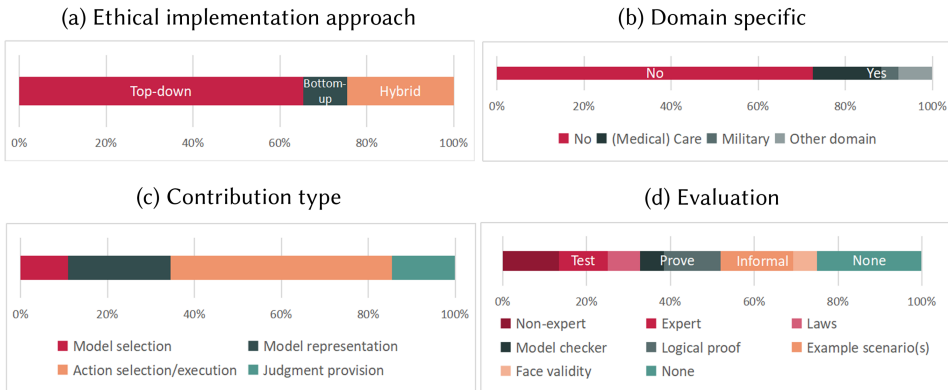

Fig. 2. Non-technical analysis.

About a quarter of the approaches are of a hybrid nature, combining at least two classical ethical theory types. Approximately half of those have a hierarchical approach, in which deontological features are standardly dominant over consequentialist ones. The non-hierarchical systems, where at least two ethical theory types work together without a single one being dominant, frequently go beyond the two main types of theory. Examples are virtue ethics and particularism [82], and a reflective equilibrium approach that combines consequences, rules, and other influences [152].

A little less than 10% of the papers do not have a specific theory implemented. Instead, they provide various proposals on how to implement different ethical theory types without choosing a particular one. This can be considered a computer scientist approach, where the goal is to devise a general framework that the users can adapt to their preferences.

It is surprising that despite previous calls that a single classical theory is not enough to create an ethical machine and hybrid methods are needed (e.g., Reference [2]), there is relatively little work on hybrid ethical machines. While most hybrid systems have emerged over the past 10 to 15 years, we could not find evidence for an increase in the creation of such systems.

## 7.2 Implementation Classification

Table 6 provides an overview of the classification of the non-technical implementation.

Approximately consistent with the number of single theory and hybrid theory approaches identified in Section 8.1, most authors choose a top-down approach. Hybrid approaches account for a little less than 25% of those chosen (see Figure 2(a)).

Most authors use a general approach to machine ethics: Almost three of four do not use a domain-specific approach but focus on a general proposal of implementing machine ethics (see Figure 2(b)).

In terms of contribution type, there is a relatively balanced division between authors investigating how an ethical machine should be shaped (model selection and model representation) and authors focusing on the output of the ethical machine (action judgment and action selection/execution, see Figure 2(c)). Most papers address action selection/execution. About 15% of all the papers focus on action judgment: The system judges a situation but leaves it up to the

Table 6. Non-technical Dimension Classification

| Appr. | Contribution type | Eval. type | Eval. subtype | Diversity | Domain | Papers |
|---|---|---|---|---|---|---|
| Top-down | Model representation | Proof | Model checker | ✓ | | Ganascia 2007 [62] |
| | | | Logical proof | | | Arkoudas et al. 2005 [13] |
| | | | | | | Bringsjord et al. 2012 [35] |
| | | | | | | Govindarajulu et al. 2017 [65] |
| | | Informal | Example scenario(s) | | | Berreby et al. 2017 [28] |
| | | None | None | ✓ | | Bonnemains et al. 2018 [32] |
| | Model selection | Informal | Example scenario(s) | ✓ | | Turilli 2007 [134] |
| | | | | ✓ | | Verheij et al. 2016 [139] |
| | | | | ✓ | | Wiegel et al. 2009 [146] |
| | Judgment provision | Test | Expert | | Engineering | McLaren 2003 [94] |
| | | | | | | Ashley et al. 1994 [15] |
| | | | Expert + Non-expert | | Military | Reed et al. 2016 [113] |
| | | Proof | Model checker | | | Dennis et al. 2015 [52] |
| | | | Logical proof | ✓ | | Mermet et al. 2016 [95] |
| | | None | None | ✓ | | Lindner et al. 2017 [89] |
| | Action selection/ execution | Test | Non-expert | | Medical | Shim et al. 2017 [126] |
| | | | | | | Dehghani et al. 2008 [50] |
| | | | Laws | ✓ | Cars | Thornton et al. 2016 [132] |
| | | | | | | Vanderelst et al. 2018 [137] |
| | | | | | | Winfield et al. 2014 [147] |
| | | Informal | Example scenario(s) | ✓ | | Cervantes et al. 2016 [38] |
| | | | | | Medical | Anderson et al. 2008 [3] |
| | | | | | | Cointe et al. 2016 [41] |
| | | | Face validity | ✓ | | Pereira et al. 2007 [102] |
| | | None | None | ✓ | | Neto et al. 2011 [100] |
| | | | | | Home care | Cloos 2005 [40] |
| | | | | | Home care | Dang et al. 2017 [135] |
| | | | | | | Anderson et al. 2004 (Jeremy) [6] |
| | Judgment provision + action selection/execution | Proof | Model checker | ✓ | | Dennis et al. 2016 [51] |
| | Model representation + action selection/execution | Informal | Face validity | ✓ | | Atkinson et al. 2008 [17] |
| Bottom-up | Model representation | Proof | Logical proof | ✓ | | Armstrong 2015 [14] |
| | | | | ✓ | | Furbach et al. 2014 [60] |
| | Model selection | None | None | ✓ | | Howard et al. 2017 [82] |
| | | | | ✓ | | Malle et al. 2017 [93] |
| | Action selection/ execution | Test | Non-expert | ✓ | | Wu et al. 2017 [151] |
| | | Informal | Example scenario(s) | | | Abel et al. 2016 [1] |
| | Model representation + action selection/execution | Test + proof | Non-expert + logical proof | ✓ | Cars | Noothigattu et al. 2018 [101] |
| Hybrid | Model representation | Test | Non-expert | | | Guarini 2006 [66] |
| | | | Expert | | | Anderson et al. 2014 [5] |
| | | None | None | | Medical | Azad-Manjiri 2014 [19] |
| | Action selection/ execution | Test | Non-expert | | | Honarvar et al. 2009 [78] |
| | | | Expert | | Medical | Anderson et al. 2006 [7] |
| | | | Laws | ✓ | Medical | Madl et al. 2015 [92] |
| | | Informal | Example scenarios | ✓ | | Yilmaz et al. 2017 [152] |
| | | | | | Military | Arkin 2007 [9] |
| | | | Face validity | | | Han et al. 2012 [72] |
| | | None | None | ✓ | | Anderson et al. 2004 (WD) [6] |
| | | | | ✓ | | Wallach et al. 2010 [144] |
| | Model selection + action selection/execution | Informal | Example scenario(s) | ✓ | | Tufis et al. 2015 [133] |
| | Judgment provision + action selection/execution | Test | Expert | ✓ | Medical | Pontier et al. 2012 [108] |

Diversity consideration: ✓ implies yes, an empty cell implies no/not present.

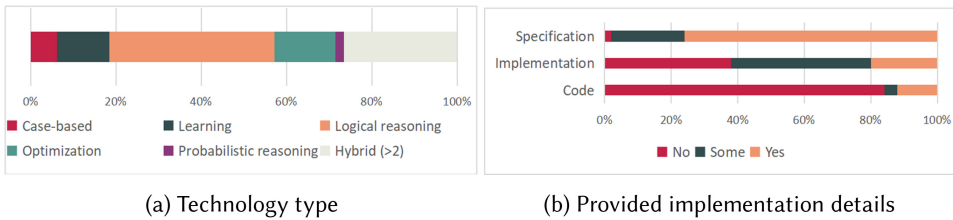

(a) Technology type                           (b) Provided implementation details

Fig. 3. Technology analysis.

human to actually act on this. From a broader scientific perspective, it is good that both model shaping and output-oriented contributions are investigated. However, it would be ideal to have both things connected.

A possibility for future improvement regards system evaluation: Over half of the authors either provide no or only an informal evaluation of their system. Of the rest, about 50% use a test approach and 50% validate their claims with some form of formal proof (see Figure 2(d)).

Finally, about half of the selected papers (51%) acknowledge diversity in implementable ethics, while the other half presents work allowing for or assuming only one ethical theory type.

### 7.3 Technical Classification

The technical dimension classification can be found in Table 7. Of the different techniques, logical reasoning is the most frequent. Figure 3(a) shows the distribution of types of technology used. About a quarter of the papers adopt more than one technology type. Only about 10% of the authors focused on a pure learning approach. Case-based reasoning and probabilistic reasoning are the least popular. Mostly classical AI approaches are used—perhaps due to the direct correspondence of rules with deontological ethics.

The level of implementation detail provided is somewhat limited (see Figure 3(b)): Although most authors include a specification of their idea in the paper, implementation details (or even source code) are rarely included. Both from a computer science perspective and a general science perspective, this is quite undesirable, as it hampers the reproducibility and extensibility of systems and empirical studies.

The different types of input used are fairly distributed: In about 36% of the ethical machines the input is defined as logical cases, in 21% the input has a numerical representation, in 30% the input is written in (natural or structured) language, and 34% use (simulated) sensor data as input. Of all cases, five selected papers had more than one type of input for their system. Around 25% of the authors used a (simulated) robot, corresponding with the amount of sensor data used as input.

In terms of user friendliness, the implemented systems score poorly. While it is important to note many of these machines are in their prototype phase and more focused on the ethics than the user, it should be important to keep the user in mind from the start of development. Nearly 35% of the machines provide an explanation of their output; 27% process the input automatically, implying that about three out of four implementations require the user to pre-process the input manually in some way—which does not make it easy for the user. Only around one out of five machines include a user interface and less than 17% offer the option for the user to give feedback. In summary, there is still plenty of room for improvement as regards user friendliness.

### 7.4 Interactions between Dimensions

Given that machine ethics is an interdisciplinary field, it is interesting to look at the interaction between the ethical theory types and their implementation, Figure 4 shows the interactions between

Table 7. Technical Classification

| Tech type | Tech subtype | Case (logical rep) | Case (numerical rep) | Case (struc) language rep) | (Simulated) Sensor data | Formalization details | Implementation details | Code (link) provided | Robot (Simulation) | Feedback | Explanation | UI | Automated Processing | Papers |
|---|---|---|---|---|---|---|---|---|---|---|---|---|---|---|
| Logical reasoning (LR) | Deductive logic | ✓ | | | | ✓ | ◦ | | | | ◦ | | | Bringsjord et al. 2012 [35] |
| | | ✓ | | | | ✓ | ◦ | | | | | | | Mermet et al. 2016 [95] |
| | | ✓ | | | | ✓ | ◦ | | | | | | | Verheij et al. 2016 [139] |
| | Non-monotonic logic (N-M logic) | ✓ | | | | ✓ | ✓ | ✓ | | | ◦ | | | Ganascia 2007 [62] |
| | Deontic logic (Deon Logic) | ✓ | | | | ✓ | ◦ | | | ◦ | ◦ | | | Arkoudas et al. 2005 [13] |
| | | ✓ | | | | ✓ | | | | ◦ | | | | Furbach et al. 2014 [60] |
| | | ✓ | | | | ✓ | | | ◦ | ◦ | | | ✓ | Malle et al. 2017 [93] |
| | | ✓ | | | | ✓ | ◦ | | | | | | | Wiegel et al. 2009 [146] |
| | Rule-based system (Rules) | | | ✓ | | ✓ | ◦ | | ✓ | | | | | Dennis et al. 2015 [52] |
| | | | | ✓ | | ✓ | ◦ | | | | | | | Dennis et al. 2016 [51] |
| | | | | ✓ | | ✓ | ◦ | | ✓ | | | | | Neto et al. 2011 [100] |
| | | | ✓ | | | ✓ | | | | | | | | Pontier et al. 2012 [108] |
| | | | | ✓ | | ✓ | ◦ | | | | | | | Tufis et al. 2015 [133] |
| | | | | | | ✓ | | | | | | | ✓ | Turilli 2007 [134] |
| | Event calculus | ✓ | | | | ✓ | | | | | | | | Bonnemains et al. 2018 [32] |
| | Abductive logic | ✓ | | | | ✓ | ✓ | | | | | | | Pereira et al. 2007 [102] |
| | N-M logic + event calculus | ✓ | | | | ✓ | ✓ | ◦ | | | ◦ | | | Berreby et al. 2017 [28] |
| | Rules + KR & ontologies | | | ✓ | | ✓ | ✓ | ✓ | | | | | | Cointe et al. 2016 [41] |
| | Deon logic + event calculus | ✓ | | | | ✓ | ✓ | ✓ | | | | | | Govindarajulu et al. 2017 [65] |
| Probabilistic reasoning (PR) | Bayes' Rule + Markov models | | | | | ✓ | ◦ | | ✓ | | | | ✓ | Cloos 2005 [40] |
| Learning (L) | Reinforcement learning | | ✓ | ✓ | | ✓ | ✓ | ✓ | | ◦ | | | | Abel et al. 2016 [1] |
| | | | | ✓ | | ✓ | ◦ | | | | | | ✓ | Wu et al. 2017 [151] |
| | Neural networks | | | ✓ | | ◦ | ◦ | | | | | | | Guarini 2006 [66] |
| | | | | ✓ | ✓ | ✓ | ◦ | | | ✓ | | | | Honarvar et al. 2009 [78] |
| | NN + Evolutionary computing | | | ✓ | | ◦ | | | ◦ | | | | | Howard et al. 2017 [82] |
| Optimization (O) | Optimization | | | ✓ | | ◦ | ◦ | | | | ✓ | ✓ | | Anderson et al. 2004 (Jeremy) [6] |
| | | | | ✓ | | ◦ | | | | | ✓ | ✓ | | Anderson et al. 2004 (WD) [6] |
| | | | ✓ | ✓ | | ✓ | | | | | | | | Anderson et al. 2008 [3] |
| | | | | ✓ | | ✓ | ◦ | | ✓ | | | | ✓ | Thornton et al. 2017 [132] |
| | | | | ✓ | | ◦ | ◦ | | | | | ✓ | | Dang et al. 2017 [135] |
| | | | | ✓ | | ✓ | ◦ | | ✓ | | | | ✓ | Vanderelst et al. 2018 [137] |
| Case-based reasoning | Case-based reasoning | ✓ | | ✓ | | ✓ | | | | ✓ | | | | Atkinson et al. 2008 [17] |
| | | ✓ | | ✓ | | ◦ | ◦ | | | ✓ | | | | Ashley et al. 1994 [15] |
| | | | | ✓ | | ✓ | ✓ | | | ✓ | | | ✓ | McLaren 2003 [94] |
| LR + L | Inductive logic | | ✓ | ✓ | | ✓ | ✓ | ✓ | | | ✓ | ✓ | | Anderson et al. 2014 [5] |
| | KR & ontologies + inductive logic | | | ✓ | | ◦ | ◦ | | | | ✓ | ✓ | | Anderson et al. 2006 [7] |
| LR + O | Deductive logic + O | | | ✓ | | ✓ | ◦ | | ✓ | | | | ✓ | Yilmaz et al. 2017 [152] |
| | Rules + O | | | ✓ | | ✓ | ◦ | | ✓ | ✓ | ✓ | ✓ | ✓ | Arkin 2007 [9] |
| | | | | ✓ | ✓ | ✓ | ✓ | ◦ | ✓ | | | | | Cervantes et al. 2016 [38] |
| | | | ✓ | | | ✓ | | | | | | | | Reed et al. 2016 [113] |
| | | | | ✓ | | ✓ | ◦ | | ✓ | | ✓ | ✓ | ✓ | Shim et al. 2017 [126] |
| | | | | ✓ | | ✓ | ◦ | | ✓ | | | | ✓ | Winfield et al. 2014 [147] |
| | Rules + abductive logic + O | ✓ | | | | ✓ | ✓ | | | ◦ | | | | Han et al. 2012 [72] |
| LR + PR | Rules + Bayes' Rule | ✓ | | | | ✓ | ◦ | ✓ | ◦ | | ◦ | | | Lindner et al. 2017 [89] |
| | Rules + statistical inference | | | ✓ | | ◦ | ◦ | | ✓ | | | ✓ | ✓ | Madl et al. 2015 [92] |
| | | | | ✓ | | ◦ | | | ✓ | ✓ | | | ✓ | Wallach et al. 2010 [144] |
| LR + CBR | Rules + KR & ontology + CBR | | | ✓ | | ◦ | ◦ | | | | ◦ | | | Dehghani et al. 2008 [50] |
| LR + L + O | Rules + decision tree + O | | ✓ | | | ✓ | | | | ✓ | | | | Azad-Manjiri 2014 [19] |
| PR + O | Bayes' Rule + O | | | ✓ | | ✓ | | | ✓ | | | | | Armstrong 2015 [14] |
| L + O | Inductive logic + O | | | ✓ | | ✓ | ◦ | | | | | | | Noothigattu et al. 2018 [101] |

✓ implies yes/fully, ◦ implies partially, an empty cell implies no/not present.

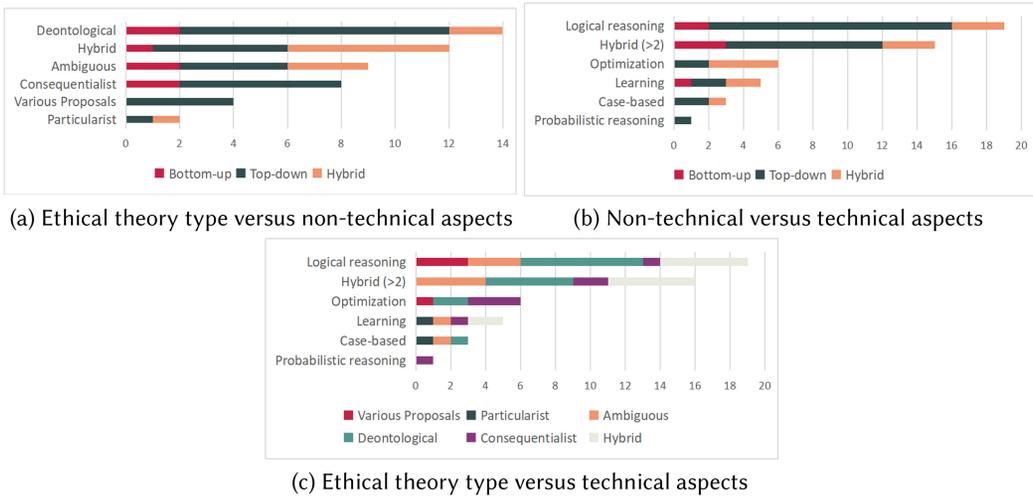

(a) Ethical theory type versus non-technical aspects

(b) Non-technical versus technical aspects

(c) Ethical theory type versus technical aspects

Fig. 4. Dimension interaction.

ethical theory, ethical implementation approach, and technology type used to implement an ethical machine. For researchers, it can be useful to see which combinations have not yet been tried out that might be promising. For example, Figure 4(a) shows that (to the best of our knowledge) a hybrid approach (including both top-down and bottom-up elements) to implementing pure consequentialism does not yet exist. Similarly, bottom-up approaches to optimization (see Figure 4(b)) or pure deontological approaches to learning (see Figure 4(c)) (e.g., seeing which input leads to behavior adherent to a certain set of rules) have not yet been explored.

## 7.5 General Observations

There are some general observations to be made about the field. First, the focus is on one universal and objective moral agent. There are barely any options for adding cultural influences or societal preferences in any of the classified papers. Almost all systems assume the user cannot influence the output of the system. A recent publication shows indication of cultural differences in ethical preferences [18], and the development of societal preferences within an ethical machine would improve the chance of acceptance of ethical machines. However, it is still under debate whether the field should move toward a "universal moral grammar," such as that proposed by Mikhail [96].

Second, there are some issues inherent to the field. For instance, there are no benchmarks to verify if a system is working as it should. There are no specific tasks to be implemented, no consensus as to what the correct output is, and few data sets to use in an implementation. A helpful tool to recur to in this context is the work by Whitley [145], who provides a four-dimensional schema for analyzing a research field. Two of the dimensions refer to the uncertainty of the task at hand, and two refer to the mutual dependence between the fields and scientists in them. The field of machine ethics scores highly on all of these dimensions:

*High technical task uncertainty*: There is unpredictability and variability in which methods are used in the field and how results are interpreted. In this regard, it is a fragmented field.

*High strategic task uncertainty*: There are problems present in the field that are valued differently (e.g., some authors focus on the theoretical, others on the implementation, and the ethical theories or even ethical theory types they focus on diverge).

*High strategic dependence*: There is much disagreement on the relevant topics, so there is a high reliance on peers for validation and reputation in the field.

*Medium functional dependence*: In terms of physical dependence of resources, there is none. Anyone with a computer can add to the field; no expensive equipment is needed. However, there is a high dependence on results of others and acceptance by the field.

Another potentially helpful perspective can be derived from Whitley's theory, where the field of machine ethics would be a "polycentric oligarchy," implying there are several independent clusters of scholars that confirm each other's assumptions and do not communicate much with other clusters that have very different views. At first glance, such clusters can indeed be detected: the multi-agent norm domain (e.g., References [100, 133]), the logical translation of ethical theories (e.g., References [65, 72, 102]), or the modern learning approach to machine ethics (e.g., References [1, 151]). While exploratory research in many directions is valuable, the field would benefit from more standardization and more communication between clusters to exchange knowledge on ethics and technology.

## 8 FUTURE AVENUES AND LIMITATIONS

Based on the results of the analysis and description of the selected papers, some literature gaps are identified that can be of interest for future work. Additionally, the limitations of this survey are discussed.

*Ethical dimension.* In view of earlier calls for hybrid systems when it comes to ethical theory, a surprisingly low percentage of authors consider a *multi-theory* approach in which machines can interchangeably apply different theories depending on the type of situation. In terms of the content (and not the structure) of ethical theories, it is important to acknowledge and harness the nuances of specific theories, but human morality is complex and cannot be captured by one single classical ethical theory. Even experts can have rational disagreement amongst themselves on an ethical dilemma. This leads to the next important point: An ethical machine will not be of use if it is not accepted by its users, which can be the risk of focusing on one ethical theory and, thus, not covering human morality. Ethical theory needs to be combined with *domain-specific ethics* as accepted by domain experts and, as identified in the analysis of this article, this is not the case in the majority of the related work. Moreover, it is necessary to discuss the ethical theory/theories in the system with its possible users. Some examples of using *folk morality* in machine ethics can be found in Noothigattu et al. [101], as well as in Reference [116]. However, it is important to note that just as ethical theories have their challenges, so does folk morality. Three challenges are who to include in the group whose values should be considered (*standing*), how to obtain their values (*measurement*), and how to aggregate their values (*aggregation*) [21]. Implementations should start from ethical theories combined with domain-specific ethical theory, after which acceptance by the users and deviation from socially accepted norms should be discussed (cf. e.g., References [18, 31, 88, 123]).

*Non-technical dimension.* There is a need for more systematic *evaluations* when ethical machines are created to be able to rate and compare systems. To this end, there is a strong need for *domain-specific benchmarks*. Based on input from domain experts, data sets need to be created containing the types of cases prevalent in that domain, with respect to which ethical machines must be assessed. The gathering of typical tasks and respective answers that domain experts agree on is just as important as the actual creation of ethical machines. This implies the need for more *collaboration* between fields. Computer scientists and philosophers, as well as domain experts and social science experts, have to work together to ensure the interaction with and effects of the ethical machines are as desired. Even within the field, collaboration is needed between different clusters of topics in the field of machine ethics, for example between clusters specializing in MAS and machine learning, respectively. Finally, in general, *implementation* requires more attention. While on a higher level, theoretical discussion remains important in this field, especially to prepare for possible future scenarios, the testing of theory in practice can enrich the discussion on what is (or is not) possible at that moment and what practical implementations and consequences certain ethical machines can have.

*Technical dimension.* When a system is implemented, it is imperative to provide exhaustive specification detail, including *availability of the code*, which is predominantly lacking. Another frequent shortcoming regards usability: The system should have a user interface so that the future user can interact with the system without having to know how to code. Furthermore, automatic processing of input cases deserves more attention, so as to avoid having to encode each variable manually as a vector for a neural network. Considering the increased need for *transparency* in algorithmic decision making, as well as the fundamental role of reasons in ethics, the system should also provide an explanation of why it took a certain decision. In a next phase, the user should be able to give *feedback* on the ethical decision the system makes. Finally, the association of a given type of technology with a certain type of ethics requires an adequate technical justification, beyond using just the most acquainted technology.

*Further Points of Interest.* Current technology allows for successful application of narrow AI geared toward specific tasks. While steps are being taken toward artificial general intelligence (AGI), the technology does not yet exist [83]. Hence, domain-specific applications seem suitable. A domain-specific non-AGI approach to machine ethics alleviates some of the risks and limitations on machine ethics posed by [36], such as those related to an "insufficient knowledge and/or computational resources for the situation at hand." However, there are still risks and limitations. For instance, in the context of lethal autonomous weapons systems, the loss of "meaningful human control" [119] is a risk, as humans would not have the same control over ethical decisions such as target selection. A limitation of using domain-specific ethical machines is that the process of one domain may not be transferable to other domains. Furthermore, not everyone is ready to accept a machine taking over the ethical decision making process [76].

A slightly different way to address ethics in machines is to define (and implement) an ethical decision support, rather than leaving the machine to make an autonomous ethical decision. For an overview of different types of moral mediation, see Van de Voort et al. [136]. Etzioni agrees that the focus should lie on decision support, stating "there seem to be very strong reasons to treat smart machines as partners, rather than as commanding a mind that allows them to function on their own" [54, p. 412]. One of those reasons is that AGI will not exist in the foreseeable future. This approach will also help with acceptance of machines with ethical considerations in society. There are different possible levels of autonomy the system can have, for example only summarizing available data, interpreting available data, summarizing possible actions, or even suggesting/preselecting a possible action the system deems best. Different types of support and collaboration might be necessary for different applications, and according to the literature review done in this article, further research is needed in this direction.

*Limitations.* This survey has some limitations that need to be mentioned. First of all, the scope of the paper selection was limited to explicit ethical theories (i.e., theories directly programmed into the machine). While some of the works reviewed can still be of interest and provide inspiration for implementation, papers devoid of implementation details were excluded from this survey. Examples are emerging ethics based on human data to research folk morality (e.g., Reference [151]) or models of human morality to determine relevant features in input cases (e.g., Reference [141]).

Furthermore, we limited the survey to one paper per author whenever similar systems were discussed across multiple publications, selecting the most comprehensive one. This does not do full justice to the work of certain authors (e.g., Guarini working on explainability of neural networks making ethical decision [67]). While the paper selection procedure was designed to be as exhaustive as possible, it is still possible that a few important papers were missed. Finally, three authors reviewed the ethical dimension and two reviewed the implementation and technical dimension, but it is still possible there was bias in the classification due to the limited number of people involved in the classification process and the process of discussion until agreement was reached.

## 9 CONCLUSION

The future of the field of machine ethics will depend on advances in both technology and ethical theory. Until new breakthroughs change the field, it is important to acknowledge what has been done so far and the avenues of research that make sense to pursue in the near future. To accomplish this, the contribution of this survey is threefold. First, a classification taxonomy with three dimensions is introduced: the ethical dimension, the dimension considering nontechnical aspects when implementing ethics into a machine, and the technical dimension. Second, an exhaustive selection of papers describing machine ethics implementations is presented, summarized, and classified according to the introduced taxonomies. Finally, based on the classification, a trend analysis is presented that leads to some recommendations on future research foci. It is important to keep in mind how machine ethics can be used in a meaningful way for its users, with increasing agreement on what a system should do, and in what context.

## APPENDIX

## A   DESCRIPTIONS OF SELECTED PAPERS

For readers that are interested in a more detailed description of the classified papers, this appendix provides a short summary of each of the selected papers. To structure their presentation, the papers were categorized across two orthogonal dimensions: (i) implementation (top-down, bottom-up, and hybrid, cf. Reference [143]), and (ii) type of ethical theory (deontological, consequentialist, virtue ethics, particularism). Given that not all dimensions for possible classification could be included to structure this section, the chosen dimensions focus on the ethical aspect of the selected papers: the ethical theory and how it is implemented.

### A.1   Top-Down

*A.1.1   Deontological Ethics.* Among top-down deontological approaches, different kinds can be distinguished: papers that use predetermined given rules for a certain domain, papers focusing on multi-agent systems (MAS), and other papers that do not fit either of these two categories.

*Domain rules.* In the medical domain, Anderson and Anderson [3] use an interpretation of the four principles of Beauchamp and Childress [22] from earlier work by Anderson et al. [7] to create an ethical eldercare system. The system, called Ethel, needs to oversee the medication intake of patients. Initial information is given by an overseer, including, for example, at what time medication should be taken, how much harm could be done by not taking the medication, and the number of hours it would take to reach this maximum harm. Shim et al. [126] also explore the medical domain, but focus on mediating between caregivers and patients with Parkinson's disease. Instead of a constraint-based approach from previous work, their paper builds on the work by Arkin [9], who employs a rule-based approach. Based on expert knowledge, a set of rules is created to improve communication quality between patient and caregiver and to ensure that the communication process is safe and not interrupted. Among other things, each rule has a type (obligation or prohibition) and response output when triggered. The rules are prohibition rules, for example about yelling, and obligations rules regarding, for instance, how to keep the patient safe. There are verbal and non-verbal cues for each action, retrieved through sensors. For the military domain, Reed et al. [113] use a model that balances the principles of civilian non-maleficence, military necessity, proportionality, and prospect of success. The resulting principles are ranked in order of importance. A scenario is used to calibrate the relative ethical violation model by updating the weight for each principle. Then, a survey is conducted to collect both expert and non-expert assessment of the situation. Rule-based systems trained on human data perform at the level of human experts. For the air traffic domain, Dennis et al. [51] developed the ETHAN system that deals with situations when civil air navigation regulations are in conflict. The system relates these rules to four hierarchical ordered ethical principles (do not harm people, do not harm animals, do not damage self, and do not damage property) and develops a course of action that generates the smallest violation to those principles in case of conflict. McLaren [94] used adjudicated cases from the National Society of Professional Engineers to adopt the principles in their code of ethics for a system called SIROCCO. Its primary goal is to test whether it can apply existing heuristic techniques to identify the principles and previous cases that are most applicable for the analysis of new cases, based on an engineering ethics ontology. SIROCCO accepts a target case in Ethics Transcription Language, searches relevant details in cases in its knowledge base in Extended Ethics Transcription Language and produces advised code provisions and relevant known cases.

*Multi-Agent Systems.* Wiegel and van den Berg [146] use a Belief-Desire-Intention (BDI) model to model agents in a MAS setting. Their approach is based on deontic epistemic action logic, which includes four steps: modelling moral information, creating a moral knowledge base, connecting moral knowledge to intentions, and including meta-level moral reasoning. Moral knowledge is linked to intentions and if there is no action that can satisfy the constraints, the agent will not act. Neto et al. [100] also implement a BDI approach for a MAS. Their focus is on norm conflict: an agent can adopt and update norms, decide which norms to activate based on the case at hand, its desires, and its intentions. Conflict between norms is solved by selecting the norm that adds most to the achievement of the agent's intentions and desires. Norm-adherence is incorporated in the agent's desires and intentions. Also, Mermet and Simon [95] deal with norm conflicts. They distinguish between moral rules and ethical rules that come into play when moral rules are in conflict. They perform a verification of whether their system called GDT4MAS, is able to choose the correct ethical rule in conflict cases.

*Other.* Bringsjord and Taylor [35] propose a normative approach using what they call "divine-command ethics." They present a divine-command logic intended to be used for lethal autonomous robots in the military domain. This logic is a natural-deduction proof theory, where input from a human can be seen as a divine command for the robot. Turilli [134] introduces the concept of the ethical consistency problem. He is interested in the ethical aspects of information technology in general. He proposes a generic two-step method that first translates ethical principles into ethical requirements, and then ethical requirements into ethical protocols.

*A.1.2 Consequentialism.* Among papers that use a top-down consequentialist approach, this survey briefly discusses (i) those that focus on the home assistance domain, (ii) those that focus on safety applications, and (iii) a variety of others.

*Home domain.* Cloos [40] proposes a service robot for the home environment. The system, called Utilibot, chooses the action with the highest expected utility. Because of the computational complexity of consequentialism, the ethical theory is a decision criterion rather than a decision process.

The description of the system seems a realistic thought experiment, mentioning features the system could have, based on previous research. The system controlling the robot, Wellnet, consists of Bayesian nets and uses a Markov decision process to optimize its behavior for its policies. Van Dang et al. [135] focus a similar use case but opt for a different technical approach: they adopt a cognitive agent software architecture called Soar. The robot is given information about family members. When it receives a request, each possible action is assigned a utility value for each general law of robotics as proposed by Asimov. The action with the maximum overall utility is selected to be executed, which can be to either obey, disobey, or partially obey (meaning proposing an alternative option for) the human's request.

*Falling prevention.* Three related papers focus on the use case where a human and robot (both represented by a robot in experiments) are navigating a space that has a hole in the ground. The robot has to decide how to intervene to prevent the human from falling into the hole.

Winfield et al. [147] add a "Safety/Ethical Logic" layer that is integrated in a so-called consequence engine, which is a simulation-based internal model. This mechanism for estimating the consequences of actions follows rules very similar to Asimov's laws of robotics. They address each law in an experiment. Dennis et al. [52] continue the work of Winfield et al. [147], by using and extending their approach, and introduce a declarative language that allows the creation of consequence engines within what they name the "agent infrastructure layer toolkit" (AIL). Systems created with AIL can be formally verified using an available model checker. The example system that is implemented sums multiple possible unethical outcomes and minimizes the number of people harmed. Vanderelst and Winfield [137] have a similar approach and implement two robots representing humans and a robot that follows Asimov's laws respectively. In their case study, there are two goal locations, one of which is dangerous, and the Asimov robot has to intervene.

*Other.* In early work by Anderson et al. [6], a simple utilitarian system is introduced based on the theory of Jeremy Bentham that implements act utilitarianism (i.e., calculates utilities of options and chooses the one with the highest utility).

*A.1.3 Particularism.* Ashley and McLaren [15] describe a system that "compares cases that contain ethical dilemmas about whether or not to tell the truth." They use a case-based reasoning approach to compare the different cases in its database. The program, called Truth-Teller, compares different real-world situations in terms of relevant similarities and distinctions in justifications for telling the truth or lying. Representations for principles and reasons, truth telling episodes, comparison rules, and important scenarios are presented.

*A.1.4 Hybrid: Specified Hierarchy.* This section contains papers that use a top-down ethical hybrid approach with a specified hierarchy. Different groups can be distinguished: papers where deontological ethics are dominant over consequentialism, and a paper where consequentialism is dominant over deontological ethics.

*Deontological dominance.* While the following three systems all have the same approach, they are very different in their implementation. In the system by Dehghani et al. [50], the ethical theory type is very clear. The system, called MoralMD, has two modes: deontological and utilitarian. A new case is processed into predicate calculus and the presence of principles and contextual features are compared to a determined set of rules in a knowledge base. The order of magnitude reasoning module calculates the relationship between the utility of each choice. If there are no sacred values involved in the case at hand (i.e., the deontological component), then the system will choose the proper output based on the highest utility (i.e., the consequentialist component). Govindarajulu and Bringsjord [65] provide a first-order modal logic to formalize the doctrine of double effect and even of triple effect: "the deontic cognitive event calculus." The calculus includes the modal operators for knowledge, beliefs, desires, and intentions. To be able to be useful in non-logic systems, they explain what characteristics a system should have to be able to use the proposed approach. The doctrine of double (and triple) effect combines deontological and consequentialist ethics, where deontology has a greater emphasis than consequentialism. Pereira and Saptawijaya [102] use prospective logical programming to model various moral dilemmas taken from the classic trolley problem and employ the principle of double effect as the moral rule. Once an action has been chosen, preferences for situations are judged a posteriori by the user. The authors show their implementation in a program called ACORDA.

*Consequentialist dominance.* In earlier work, Pontier and Hoorn [108] introduced a "cognitive model of emotional intelligence and affective decision making" called Silicon Coppélia to be used in the health domain. An agent has three moral duties (autonomy, beneficence, and non-maleficence) with a certain ambition to fulfill each duty (i.e., weights). The system's decisions are based on action-specific expected utilities and consistency with the predetermined duties. While most authors make an act utilitarian system, Pontier and Hoorn create a rule utilitarian system by trying to maximize the total amount of utility for everyone. While they use rules (i.e., deontological ethics), they implement them in a consequentialist way, making this the dominant ethical theory type. Their model was extended to match decisions of judges in medical ethical cases [109].

*A.1.5 Hybrid: Unspecified Hierarchy.* Both systems in this category focus on a modular approach, where different ethical theory types can be combined in an ethical machine. The goal of the system by Berreby et al. [28] is to create a modular architecture to represent ethical principles in a consistent and adjustable manner. They qualify what they call "the Good" and "the Right" as the ethical part of their system (implying both consequentialist and deontological constraints). Besides these system components, the system consists of an action model (i.e., "it enables the agent to represent its environment and the changes that take place in it, taking as input a set of performed actions") and a causal model (i.e., "it tracks the causal powers of actions, enabling reasoning over agent responsibility and accountability, taking as input the event trace given by the action model and a specification of events containing a set of events and of dependence relations") [28]. The implementation is done in Answer Set Programming using a modified version of Event calculus. Using a medical scenario, they provide a proof of concept. Lindner et al. [89] have created a software library for modelling "hybrid ethical reasoning agents" called HERA. Based on logic, they create a prototype called IMMANUEL, which is a robotic face and upper body that users can interact with. The system's ethical constraints draw on consequentialist calculations, the Pareto principle from economics, and the principle of double effect. Uncertainty and belief in permissibly of an action are added as extra variables in the system.

*A.1.6 Configurable Ethics.* The papers in this subsection have a top-down approach and proposed various ways in which ethics can be implemented. One paper has machine ethics tailored for a specific domain, while another uses different techniques in a more domain-general way. A third focuses on multi-agent systems.

*Domain-specific.* Thornton et al. [132] combine deontology, consequentialism, and virtue ethics to optimize driving goals in automated vehicle control. Constraints and costs on vehicle goals are determined on the basis of both deontological and consequentialist considerations. Virtue ethics generates specific goals across vehicle types, such that a traffic infraction of an ambulance is assessed as less costly than that of a taxi cab.

*Domain-general.* Ganascia [62] claims to be the first to attempt to model ethical rules with Answer Set Programming (cf. [20]) to model three types of ethical systems—Aristotelian ethics,

Kantian deontology, and Constant's "Principles of Politics" (cf. [43]). Drawing on [107] situation calculus, Bonnemains et al. [32] devise a formalism in which moral dilemmas can be expressed and resolved in line with distinct ethical systems, including consequentialism and deontological ethics.

*Multi-agent systems.* Cointe et al. [41] extend ethical decision making to multi-agent systems. The judgment function can accommodate a wide variety of inputs and is not restricted to the format of a single type of ethical system.

*A.1.7 Ambiguous.* Arkoudas et al. [13] reason that well-behaved robots should be based on "mechanized formal logics of action, obligation and permissibility." After introducing a domain-specific deontic logic, they describe a previously published interactive theorem proving system, Athena, that can be utilized to verify ethical systems based on first-order logic. Murakami [99] presented an axiomatization of Horty's utilitarian formulation of multi-agent deontic logic [80], while Arkoudas et al. [13] present a sequent-based deduction formulation of Murakami's system. While deontic logic is used, each deontic stit frame contains a utility function. The contribution lies in the new approach to Murakami's system, which is implemented and proven in Athena. In a different approach, the proposed system by Cervantes et al. [38] devise a computational model for moral decision-making inspired by neuronal mechanisms of the human brain. The model integrates agential preferences, past experience, current emotional states, a set of ethical rules, as well as certain utilitarian and deontological doctrines as desiderata for the impending ethical decision.

With an entirely different focus, Atkinson and Bench-Capon [17] depart from Hare's contention [73] that in situations with serious consequences, we engage in complex moral reasoning rather than the simple application of moral rules and norms. Moral norms are thus considered not an input to, but an output of serious moral deliberation. The authors model situated moral reasoning drawing on *Action-Based Alternating Transition Systems* (cf. Reference [148] as well as Reference [16]). While some argue this approach can be seen as virtue ethics (e.g., [23]), the authors of this survey consider this to be a consequentialist implementation, as the focus of the approach is on whether the consequences of an action adhere to a certain value.

Verheij [139] draws on Bench-Capon's framework of value-based argumentation [24, 25], which is inspired by case law (new cases are decided on past cases where there is no clear legislation, cf. Reference [70]). The paper, focusing on computational argumentation for AI in Law, breaks new ground in so far as the formal model is not restricted to either qualitative or quantitative primitives, but integrates both.

## A.2 Bottom-up

*A.2.1 Deontological Ethics.* Malle et al. [93] argue that robots need to have a norm capacity—a capacity to learn and adhere to norms. Drawing on deontic logic, the authors explore two distinct approaches of implementing a norm system in an artificial cognitive architecture. Noothigattu et al. [101] collect data on human ethical decision making to learn societal preferences. They then create a system that summarizes and aggregates the results to make ethical decisions.

*A.2.2 Consequentialism.* Armstrong [14] observes that equipping artificial agents directly with values or preferences can be dangerous (cf. Reference [33]). Representing values as utility functions, the author proposes a value selection mechanism where existing values do not interfere with the adoption of new ones. Abel et al. [1] pursue a related goal. In contrast to Armstrong, the agent does not maximize a changing meta-utility function but instead draws on partially observable Markov decision processes (cf. Reference [85]) familiar from reinforcement learning. The system is tested with respect to two moral dilemmas.

*A.2.3 Hybrid: Unspecified Hierarchy.* In contrast to the dominant action-based models of autonomous artificial moral agents, Howard and Muntean [82] advocate an agent-based model, which combines traits of virtue ethics and moral particularism. The implementation draws on neural networks optimized by evolutionary computation and is given a test run with the NEAT (NeuroEvolution of Augmenting Topologies) package (cf. References [55, 114, 129, 130]).

*A.2.4 Ambiguous.* Furbach et al. [60] demonstrate how deontic logic can be transformed into description logic so as to be processed by Hyper—a theorem prover employing hypertableau calculus by aid of which normative systems can be evaluated and checked for consistency. Wu and Lin [151] are interested in "ethics shaping" and propose a reinforcement learning model. The latter is augmented by a system of penalties and rewards that draws on the Kullback-Leibler divergence [87].

## A.3 Hybrid

This section introduces selected papers that use a hybrid approach to implement ethics by combining top-down and bottom-up elements.

*A.3.1 Deontological Ethics.* The following papers, all by the same set of authors, use a hybrid approach to implement deontological ethics. In 2004, Anderson et al. [6] introduced *W.D.*, a system based on the *prima facie* duties advocated by W. D. Ross. W.D. leaves the encoding of a situation up to the user, who has to attribute values to the satisfaction and violation of the duties for each possible action. The system pursues the action with the highest weighted sum of duty satisfaction. Two years later, Anderson et al. [7] introduced *MedEthEx*, an advisory system in medical ethics. MedEthEx has three components: a basic module trained by experts, a knowledge-based interface that guides users when inputting a new case, and a module that provides advice for the new case at hand. In 2014, Anderson and Anderson [5] created *GenEth*, a general analyzing system for moral dilemmas. The system is capable of representing a variety of aspects of dilemmas (situational features, duties, actions, cases, and principles) and can generate abstract ethical principles by applying inductive logic to solutions of particular dilemma cases. The principles are evaluated by a self-made Ethical Turing Test: If the system performs as an ethical expert would, then it passes the test. GenEth was also applied in a eldercare use case [8].

*A.3.2 Particularism.* Guarini [66] explores whether neural networks can be employed to implement particularist ethics, as occasionally hinted at by Dancy, one of particularism's most renowned advocates (cf. References [46–48]). Using the action/omission distinction (cf. Reference [150] for a review) as a test paradigm, neural networks are trained with different types of cases to investigate whether they can competently judge new ones.

*A.3.3 Hybrid: Specified Hierarchy.* Arkin [9] explores constraints on the deployment of lethal autonomous weapons in the battlefield (it was subsequently published as a series of three articles [10–12]. The proposed system is predominantly governed by deontological rules, namely international laws of war and the U.S. Army's rules of engagement. Its architecture relies on four central constituents: an Ethical Governor that suppresses lethal action; an Ethical Behavior Control that constrains behavior in line with the rules; an Ethical Adaptor, which can update the agent's constraint set to a more restrictive one; and a Responsibility Advisor, which is the human-robot interaction part of the system.

Azad-Manjiri [19] develops an architecture for a deontological system constrained by Beauchamp and Childress's biomedical principles. The system determines its actions on the basis of said principles and a decision tree algorithm trained with expert ethicist judgments in a variety of cases from the biomedical domain. Building on early work by Ganascia [63], Tufiş and Ganascia [133] augment a belief-desire-intention rational agent model with normative constraints. They devote particular attention to the problem arising from the acquisition of new norms, which frequently stand in conflict with existing ones (for an alternative approach building on the belief-desire-intention model, see Honarvar and Ghasem-Aghaee [78, 79] discussed below).

*A.3.4   Hybrid: Unspecified Hierarchy.* Yilmaz et al. [152] survey the field of machine ethics and propose a coherence-driven reflective equilibrium model (cf. Reference [112]), by aid of which conflicts across heterogenous interests and values can be resolved. Honarvar and Ghasem-Aghaee [78] build a belief-desire-intention agent model whose decisions are based on a number of weighted features drawn from hedonic act utilitarianism (e.g., the amount of pleasure and displeasure for the agent and other parties affected by the action).

*A.3.5   Ambiguous.* Most of the work of Saptawijaya and Pereira (cf. References [102–105, 120–122]) focuses on logic programming and prospective logic to model ethical machines. In Han et al. [72], they introduce uncertainty as a factor in decision making and draw on abductive logic to accommodate it. Madl and Franklin [92] call for limits on ethical machines for safety reasons. Developing on Franklin et al.'s [59] LIDA architecture—an AGI model of human cognition—they suggest that deliberate actions could be constrained top-down during run time, and ethical meta-rules (such as certain Kantian principles) could be implemented on a metacognitive level. Rather than start from a complete set of rules, the latter can gradually expand. The approach is exemplified by *CareBot*, an assistive simulated bot for the home care domain. Wallach et al. [144] also discuss the LIDA model. They demonstrate how emotions can be integrated into a LIDA-based account of the human decision making process and extend the approach to artificial moral agents.

## ACKNOWLEDGMENTS

We thank the authors of the selected papers for providing valuable feedback on their paper's representation. Also, we thank the anonymous reviewers for their feedback on our manuscript, which has helped us to substantially improve it.